\documentclass[letterpaper]{article} 
\usepackage{aaai25}  
\usepackage{times}  
\usepackage{helvet}  
\usepackage{courier}  
\usepackage[hyphens]{url}  
\usepackage{graphicx} 
\usepackage{natbib}  
\usepackage{caption} 
\usepackage{algorithm}
\usepackage{algorithmic}
\usepackage{subfigure}
\usepackage{amsmath}
\usepackage{url}
\usepackage{booktabs}
\usepackage{color}
\usepackage{array}
    
\newcommand{\eg}{\textit{e}.\textit{g}., }
\newcommand{\ie}{\textit{i}.\textit{e}., }
\usepackage{newfloat}
\usepackage{listings}
\usepackage{cleveref}
\DeclareCaptionStyle{ruled}{labelfont=normalfont,labelsep=colon,strut=off} 
\floatstyle{ruled}
\newfloat{listing}{tb}{lst}{}
\floatname{listing}{Listing}
\nocopyright

\title{Hybrid SD: Edge-Cloud Collaborative Inference for Stable Diffusion Models}
\author{
    Chenqian Yan\equalcontrib, Songwei Liu\equalcontrib, Hongjian Liu\equalcontrib,  Xurui Peng, \\ Xiaojian Wang, 
    Fangmin Chen, Lean Fu, Xing Mei\\
}
\affiliations{
    ByteDance Inc.

}

\usepackage{bibentry}
\usepackage{multirow}

\usepackage{amsmath,amsfonts,bm}

\def\eqref#1{equation~\ref{#1}}

\def\1{\bm{1}}

\def\vepsilon{{\bm{\epsilon}}}

\def\vtheta{{\bm{\theta}}}

\def\vc{{\bm{c}}}

\def\vx{{\bm{x}}}

\def\vz{{\bm{z}}}

\DeclareMathAlphabet{\mathsfit}{\encodingdefault}{\sfdefault}{m}{sl}
\SetMathAlphabet{\mathsfit}{bold}{\encodingdefault}{\sfdefault}{bx}{n}

\begin{document}

\maketitle

\begin{abstract}
  Stable Diffusion Models (SDMs) have shown remarkable proficiency in image synthesis. However, their broad application is impeded by their large model sizes and intensive computational requirements, which typically require expensive cloud servers for deployment.
  On the flip side, while there are many compact models tailored for edge devices that can reduce these demands, they often compromise on semantic integrity and visual quality when compared to full-sized SDMs.
  To bridge this gap, we introduce Hybrid SD, an innovative, training-free SDMs inference framework designed for edge-cloud collaborative inference.
  Hybrid SD distributes the early steps of the diffusion process to the large models deployed on cloud servers, enhancing semantic planning. Furthermore, small efficient models deployed on edge devices can be integrated for refining visual details in the later stages.
  %
  %
  %
  %
  %
  Acknowledging the diversity of edge devices with differing computational and storage capacities, we employ structural pruning to the SDMs U-Net and train a lightweight VAE. 
  Empirical evaluations demonstrate that our compressed models achieve state-of-the-art parameter efficiency (225.8M) on edge devices with competitive image quality. Additionally, Hybrid SD reduces the cloud cost by 66\% with edge-cloud collaborative inference.


\end{abstract}

%

\section{Introduction}
Stable Diffusion Models (SDMs) \cite{Rombach_2022_CVPR, podell2023sdxl} have emerged as a pivotal technique in image synthesis, primarily due to their outstanding capability in synthesizing diverse and high-quality content. 
The remarkable generative capabilities have driven SDMs as a backbone in various generative applications, including super-resolution \cite{LI202247}, image editing \cite{kawar2023imagic, hou2024high},  text-to-image \cite{Rombach_2022_CVPR, saharia2022photorealistic, zhang2023adding}, text-to-video~\cite{peng2024conditionvideo}. The high performance and superior generative quality of SDMs always come at the expense of larger model size and more computational overheads,  \eg SDXL \cite{podell2023sdxl} base model has 3.5 billion parameters, and rectified flow model \cite{esser2024scaling} has 8 billion parameters. These models impose immense computational demands, often necessitating cloud-based inference implementations with high-end GPUs. 
However, deploying SDMs on the cloud brings high costs and potential privacy concerns, especially in scenarios where private images and prompts are uploaded to the third-party cloud service.

\begin{figure}[t]
    \centering
    \includegraphics[width=1.0\linewidth]{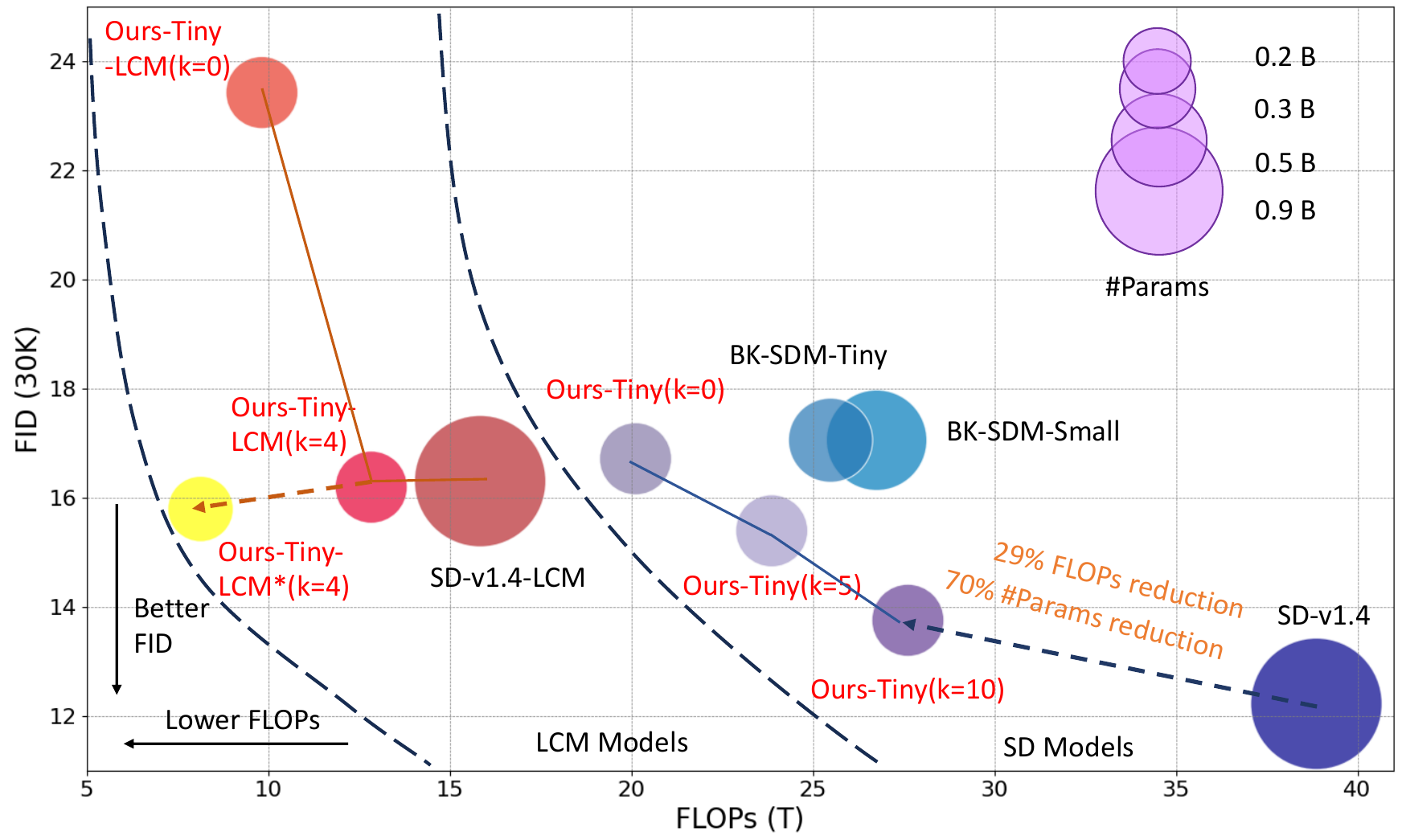}
    \caption{Comparisons of FLOPs, model size (\#Params) and FID score on MS-COCO 2014 30K dataset~\cite{lin2014microsoft}. 
    We report FLOPs and parameters of the U-Net and VAE decoder for each model.
    Our proposed Hybrid SD is highlighted in red font, where $k$ indicates the number of steps running on cloud servers. For all the SDMs, we deploy a 25-step DPMSolver~\cite{lu2022dpm} sampler. Hybrid SD achieves the compelling FID with minimal parameters and computational costs. 
    The region between the two dashed lines represents the accelerated LCM models with 8-step sampling by default. Hybrid SD shows exceptional compatibility with accelerated models. * represents replacing the original VAE with our lightweight VAE on edge devices.
    }
    \vspace{-0.05in}
    \label{fig:systems_vae}
\end{figure}

The privacy concerns and the high computational costs associated with cloud inference, particularly given the increasing number of daily active users, have sparked interest in on-device SDMs. Previous methods including efficient structure evolving \cite{li2023snapfusion}, structural pruning \cite{castells2024ldpruner, kim2023architectura}, and quantization \cite{li2023qdiffusion}, have demonstrated the feasibility of running SDMs on edge devices. However, 
empirical evaluations in \cite{li2023snapfusion} as well as ours in Figure \ref{fig:systems_vae}  show that lightweight models typically lag behind the full-sized SDMs, especially in terms of generative quality and semantic text-image alignment.

\begin{figure}[t]
    \centering
    \includegraphics[width=1.\linewidth]{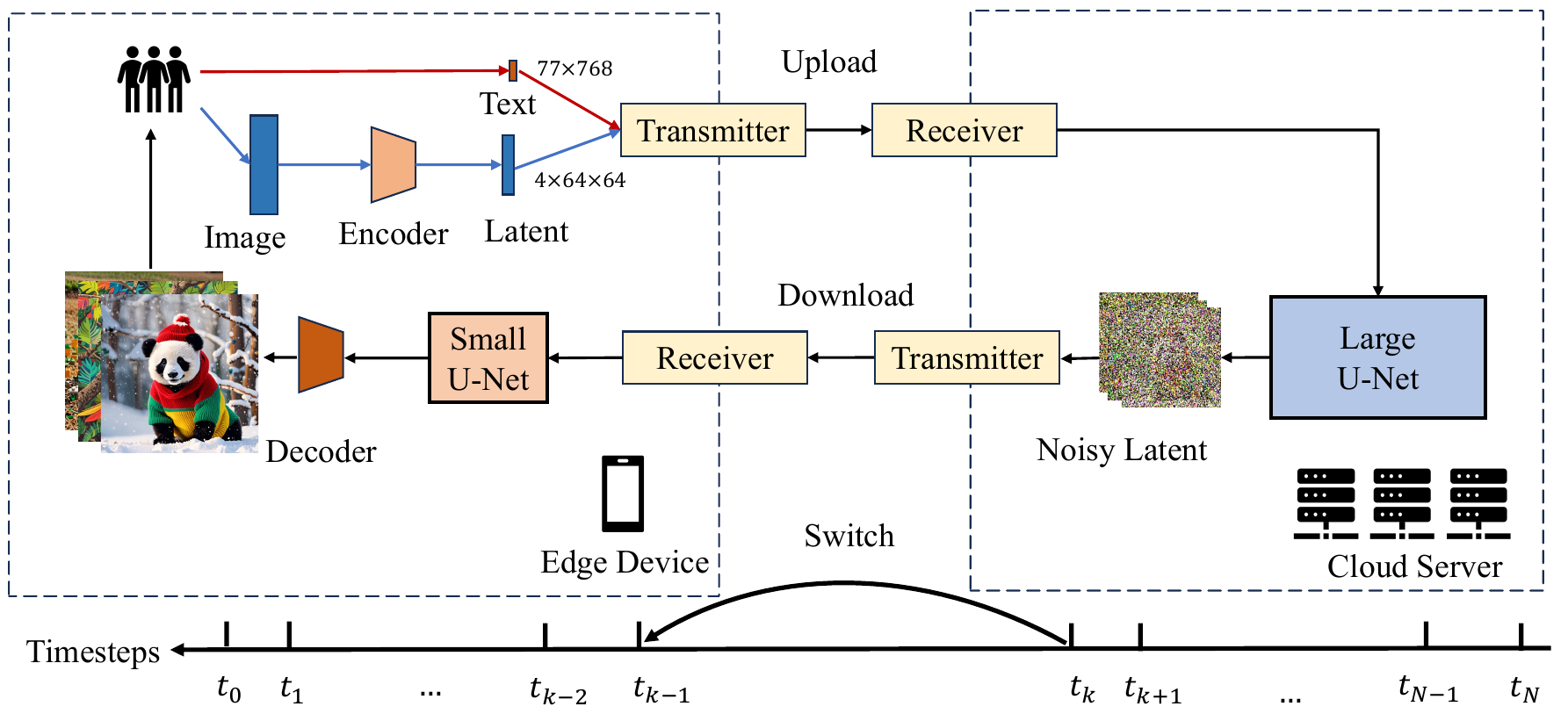}
    \caption{ The overview of Hybrid SD. We distribute the inference tasks to cloud servers and edge devices.
    The red line denotes text-to-image tasks while the blue line denotes image-to-image tasks.}
    \label{fig:systems}
    \vspace{-1em}
\end{figure}

In this work, we propose the first edge-cloud collaborative SDMs inference paradigm termed ``Hybrid SD". Figure \ref{fig:systems} gives an overview of our Hybrid SD framework. 
Aiming to shift specific computational tasks from cloud to edge, we strategically distribute the inference to large cloud-based models and small edge-based models. The large cloud-based models exhibit enhanced capabilities when it comes to planning visual semantics that are oriented toward textual content. It plays a pivotal role in the initial phases of the denoising process, where the foundational structure and semantic clarity are established. Conversely, the small edge model deployed on edge devices, while less adept at integrating semantic information, is well-suited for the later stages of the denoising process. In these stages, the focus shifts towards the recovery and enhancement of perceptual information, where the smaller model's efficiency can be effectively utilized to refine the visual details and ensure that the final images are perceptually coherent and semantically aligned with the textual input.



Concurrently, 
to further alleviate the model size and computational pressure on the edge side, we propose a smaller UNet model and an improved VAE model.
Compared with the original SD1.4, we achieve an unprecedented reduction in edge device SDMs (909.0M v.s. 225.7M), while maintaining a competitive Frechet Inception Distance (FID)~\cite{heusel2018gans} (12.22 v.s. 13.75) within our proposed hybrid inference framework. Additionally, we extend Hybrid SD to step-distillation methods (\eg LCM~\cite{luo2023latent}) and further show its compatibility.

The main contributions of this paper are as follows:
\begin{itemize}
\item We propose a novel edge-cloud collaborative inference strategy for stable diffusion models, called Hybrid SD, which avoids directly uploading user data to the cloud while reducing the cloud inference cost by 66\%.

\item To meet the restrictions of edge devices, we employ structural pruning in U-Nets and train a lightweight VAE. Our on-device models achieve state-of-the-art parameter efficiency with compelling visual quality.
\item Extensive experiments demonstrate that our approach excels in striking an optimal balance between performance and efficiency.

\end{itemize}

\section{Related Work}

\textbf{Diffusion Model Acceleration.} 
The practical application of diffusion models is hindered by their expensive computational cost and slow iterative sampling process during inference. There are two major approaches aiming to solve these problems. Solver-based methods discretize the diffusion process and explore training-free ordinary differential equation (ODE) solvers~\cite{song2020denoising, lu2022dpm}
to reduce the number of iteration steps required for the inference of diffusion models. However, they fail to generate satisfactory samples within a few steps (\eg $4\sim10$).
Distillation-based methods~\cite{salimans2022progressive, meng2023distillation, luo2023latent, lin2024sdxllightning} progressively transfers knowledge from a pre-trained teacher model to a fewer-step student model with the same architecture. These methods achieve impressive results within few-step inference. 
Despite these methods accelerating the inference of diffusion models, the student model adopts the same architecture of the teacher model, which requires a lot of memory and computational resources, prohibiting their application on edge devices. \\
\hspace{-1em}\textbf{Compression of Diffusion Model.} 
To compress the diffusion model, previous techniques can be divided into several categories: architecture redesign \cite{yang2023diffusion}, network pruning  \cite{fang2023structural, castells2024ldpruner}, quantization \cite{li2023qdiffusion}. We focus on structural pruning of diffusion models, which aims to remove structural weights, including convolution filters or linear features. There are several pruning units in structural pruning, some works adopt the entire block removing \cite{kim2023architectura}, which efficiently removes a large amount of parameters, but is hard to control the model size. Others take the operators \cite{castells2024ldpruner} that remove some computation based on the evaluation score, which do not take the number of the parameters into account. The work in \cite{fang2023structural} prunes parameters from the perspective of filters, aligning with our objective. However, it uniformly applies the same pruning ratio across all layers, overlooking the varying significance of each layer. Efforts have been initiated to develop lightweight diffusion models~\cite{li2023snapfusion}, yet these endeavors have not guaranteed semantic consistency on par with their larger counterparts.\\



\vspace{-1em}
\hspace{-1em}\textbf{Hybrid Inference.} Several works aim to combine diffusion models with different sizes for inference. 
Ediff-i~\cite{balaji2022ediff} incorporates multiple expert models to enhance the output quality, neglecting efficiency. OMS-DPM~\cite{liu2023oms}  proposes a model schedule to select different models for different sampling steps.
DDSM~\cite{yangdenoising} trains variable-sized neural networks for different steps of the diffusion process. However, it
remains uncertain whether the training strategy of the variable-sized network can effectively and robustly scale up to larger and more complex models \eg SDXL. 
Additionally, the variable-sized network does not reduce the number of parameters in the model, which hinders its deployment on edge devices.
Our work distinguishes these methods in an edge-cloud collaborative manner. For collaborative inference, previous works integrate the resources of edge devices and cloud servers for efficient inference. 
\cite{Distributed} maps parts of a model onto distributed devices. Other works~\cite{Ren_2023, liu2020datamix} mainly focused on traditional small models \eg 
VGG, ResNets. Considering the much larger model size and iterative sampling nature of diffusion models, it is non-trivial to directly adopt these methods. 
Recently, \cite{tian2024edge} proposes an edge-cloud collaborative framework. However, they focus on general distributed training and system service design, whereas our approach is specifically tailored to enable collaborative inference for SDMs. 
The application of collaborative inference to SDMs introduces unique challenges that demand a reevaluation of conventional strategies.

\section{Preliminary}
\textbf{Diffusion Model Objectives.} 
Given a data distribution $\vx_0 \sim q(\vx_0)$, the forward diffusion process progressively add Gaussion noise to the $\vx_{t-1}$ as follows,

\begin{equation}
\label{eq:forward_process}
q\left(\mathbf{x}_{t} \mid \mathbf{x}_{t-1}\right)=\mathcal{N}\left(\mathbf{x}_{t} ; \sqrt{1-\beta_{t}} \mathbf{x}_{t-1}, \beta_{t} \mathbf{I}\right)
\end{equation}

 \noindent where $t$ is the current timestep, T denotes the set of the timesteps, $\vx_1, ...\vx_T$ refer to a sequence of noisy latent, $\beta_{t}$ is a pre-defined variance schedule which describes the amount of noise added at each timestep $t$, $\mathbf{I}$ is the identity matrix with the same dimensions as the input $\vx_0$, and $\mathcal{N}(\vx;\mu,\sigma)$ means the normal distribution with mean $\mu$ and covariance $\sigma$.
The reverse diffusion process denoises the observation $\vx_t$ to estimate $\vx_{t-1}$. This process is approximated by training a noise predictor $\vtheta$ for all timesteps to learn a data distribution $p_\vtheta(\vx)$
\begin{equation}
\label{eq:backward_process}
\begin{array}{l}
p_{\vtheta}\left(\vx_{t-1} \mid \vx_{t}\right)=
\mathcal{N}\left(\vx_{t-1} ; \mu_{\vtheta}(\vx_t, t), \Sigma_{\vtheta}(\vx_t, t)\right)
\end{array}
\end{equation}

 \noindent where $\mu_{\vtheta}(\cdot,\cdot)$ and $\Sigma_{\vtheta}(\cdot, \cdot)$ are the trainable mean and covariance functions, respectively.

For stable diffusion models~\cite{Rombach_2022_CVPR}, the diffusion process is applied in the latent space of a pre-trained variational autoencoder (VAE), where an image encoder $\mathcal{E}$, an image decoder $\mathcal{D}$ and conditioning $c$ are introduced. The noise predictor $\vtheta$ is trained on the objective.

\begin{equation}
\label{eq:learning_objective}
    \mathcal{L_{\vtheta}}:=\mathbb{E}_{\vz \sim \mathcal{E}(\vx), \vc, t, \epsilon \sim \mathcal{N}(0, \mathbf{I})}\left[\left\|\vepsilon-\vepsilon_{\vtheta}\left(\vz_{t}, t, c\right)\right\|_{2}^{2}\right]
\end{equation}

\noindent\textbf{Prunable Structure.} U-Net, as the predominant conditional noise predictor and the core subject of our study, comprises primarily of ResNet blocks and Spatial Transformer blocks. In detail, each ResNet block encompasses a pair of $3 \times 3$ convolutions layers with identical filter counts. A Spatial Transformer block integrates a Self-Attention layer followed by a Cross-Attention layer. 
To uphold architectural integrity and avoid mismatches in channel configurations, our objective is to preserve the output channels of these fundamental blocks intact. 
With a view to achieving this, we target the first convolutional layer within ResNet blocks for pruning, leading to a decrease in input channels for the successive convolutional layer without disrupting the block's output structure.
Similarly, we adopt pruning of attention heads in both Self-Attention and Cross-Attention layers within the Spatial Transformer blocks, thereby efficiently adjusting the model's complexity without compromising the alignment of feature channels. This strategic pruning approach ensures compatibility and maintains the functional coherence of the network, even after significant size reduction.

\begin{figure}[t]
    \centering
    \includegraphics[width=1.0\linewidth]{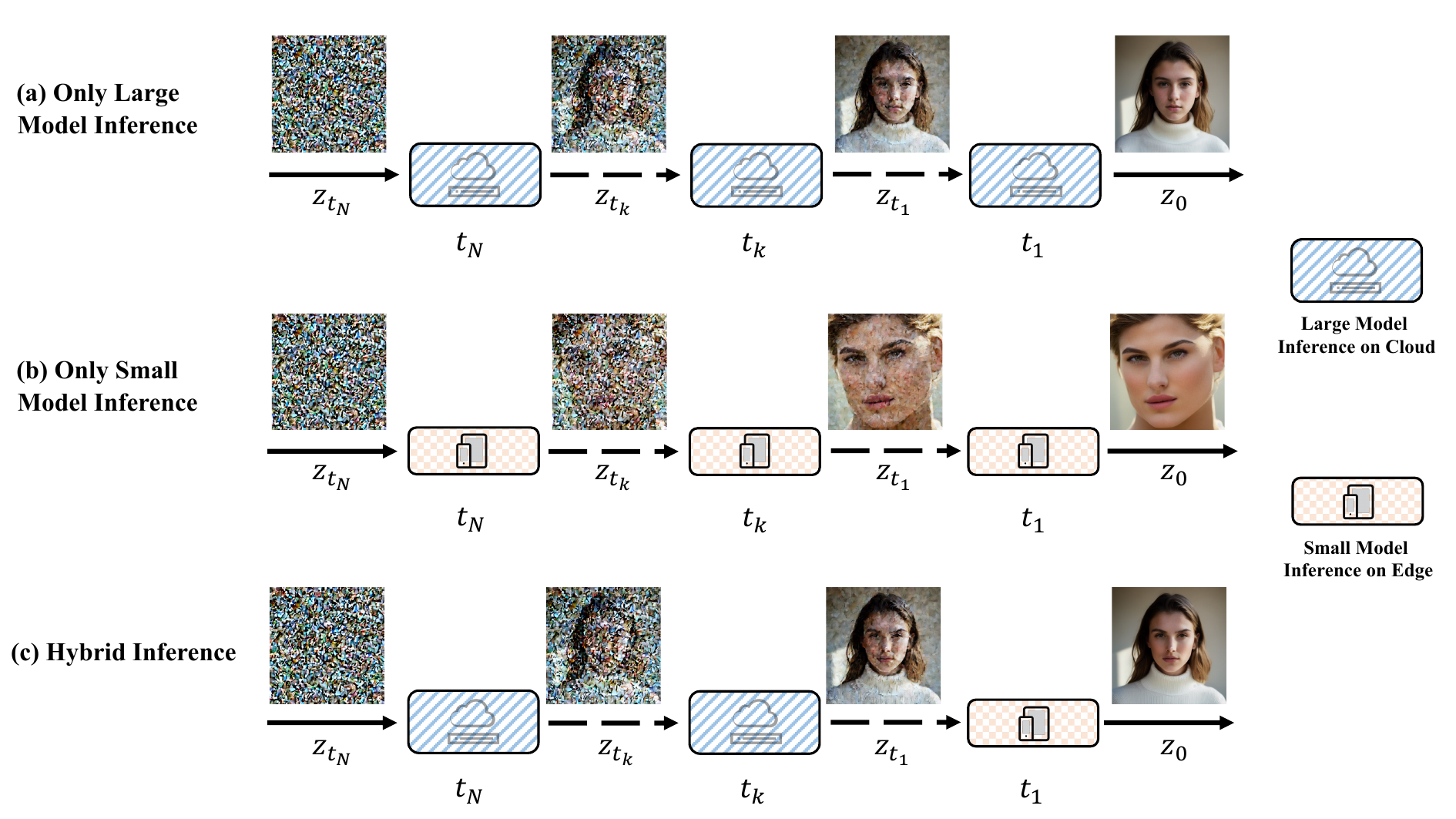} 
    \caption{Illustration of different SDMs inference process. (a) Large SD model inference on cloud. (b) Small SD model inference on edge. (c) Hybrid SD inference in a edge-cloud collaborative manner.
}
    \label{fig:framework}
\end{figure}

\section{Hybrid SD}
In this section, we present Hybrid SD for edge-cloud collaborative inference. We motivate our method by~\cite{tgate}, which characterizes the denoising steps by semantics-planning and fidelity-improving stages. The semantics-planning stage embeds text through cross-attention to obtain visual semantics. The fidelity-improving stage improves the generation quality without the requirement of cross-attention. This indicates that in the early stage of denoising, the noise predictor plays an important role in encoding conditioning information into the image latent while in the later steps, the noise predictor mainly focuses on recovering the visual perception information.

\subsection{Hybrid Inference}
Instead of conventional approaches that rely on a single model for denoising, we employ a hybrid inference strategy that integrates two distinct models for denoising. The first is a large model $\vtheta_{large}$, which is deployed in the cloud, and the second is a small and compact model $\vtheta_{small}$, deployed on edge devices. Figure \ref{fig:framework} illustrates different diffusion pipelines. We redefine the reverse diffusion process, initially outlined in Eq. \ref{eq:backward_process} to accommodate our hybrid methodology as follows:
\begin{equation}
\begin{aligned}
\label{eq:moidfied_backward_process}
\mspace{-20mu} p_{M(t, k)}&\left(\vz_{t-1} \mid \vz_{t}\right)=\\
&\mathcal{N}\left(\vz_{t-1} ; \mu_{M(t, k)}(\vz_t, t), \Sigma_{M(t, k)}(\vz_t, t)\right)
\end{aligned}
\end{equation}
\noindent where $M(t, k)$ returns different models according to current step $t$ and a pre-defined split step $k$. For steps where $t > k$, $\vtheta_{large}$ is used for denoising, otherwise, $\vtheta_{small}$ is employed for the subsequent steps. Our framework encompasses the traditional single-model approach as a special scenario, where $M(t,k)$ consistently selects the same model throughout the process. We provide the pseudocode in Appendix \ref{alg: hybrid}. Our approach leverages the complementary strengths of two models with different sizes in the reverse diffusion process, effectively minimizing inference expenses without compromising the fidelity of the synthesis.

\noindent\textbf{Analysis of cost and efficiency.}
Firstly, we delve into the key advantage of our proposed Hybrid SD inference paradigm: substantial cost reduction.
%
 Deploying the standard SDXL model via AWS services for 8 hours daily over 20 working days equates to a monthly expense of \$310 
 \footnote{\url{awslabs.github.io/stable-diffusion-aws-extension/en/cost}}.
Assuming ten applications go live each month, with each application deployment requiring 3,000 models, if half of the inference tasks are offloaded to user terminals for processing, this would result in an annual savings of 50 million.
\noindent Another concern would be the communication latency between the cloud and edge. 
Let $t$ be the transmission time, $D$ denote the data size, and $B$ be the network bandwidth, we have $t = \frac{d}{B}$. We leverage the mean WiFi bandwidth of 18.88Mbps reported at \cite{8737614}. Consider the baseline where all the inference of SD-v1.4~\cite{hf-sdv1-4} is deployed on the cloud and an image sized $3\times512 \times 512$ (768KB in 8-bit precision) is sent to edge devices, the transmission time is 0.333s.
While in Hybrid SD, two key data are transferred from the cloud to the edge: noise latent sized at $4 \times 64 \times 64$ and text embeddings sized at $77\times768$. The cumulative data in FP16 precision is totals 148KB. The cost of transmitting 148KB data is approximately 0.064s, 
posing a smaller addition to the diffusion's overall latency.

\subsection{Smaller Models}
\begin{figure}[t]
    \centering
    \subfigure[]{\includegraphics[width=0.32\columnwidth]{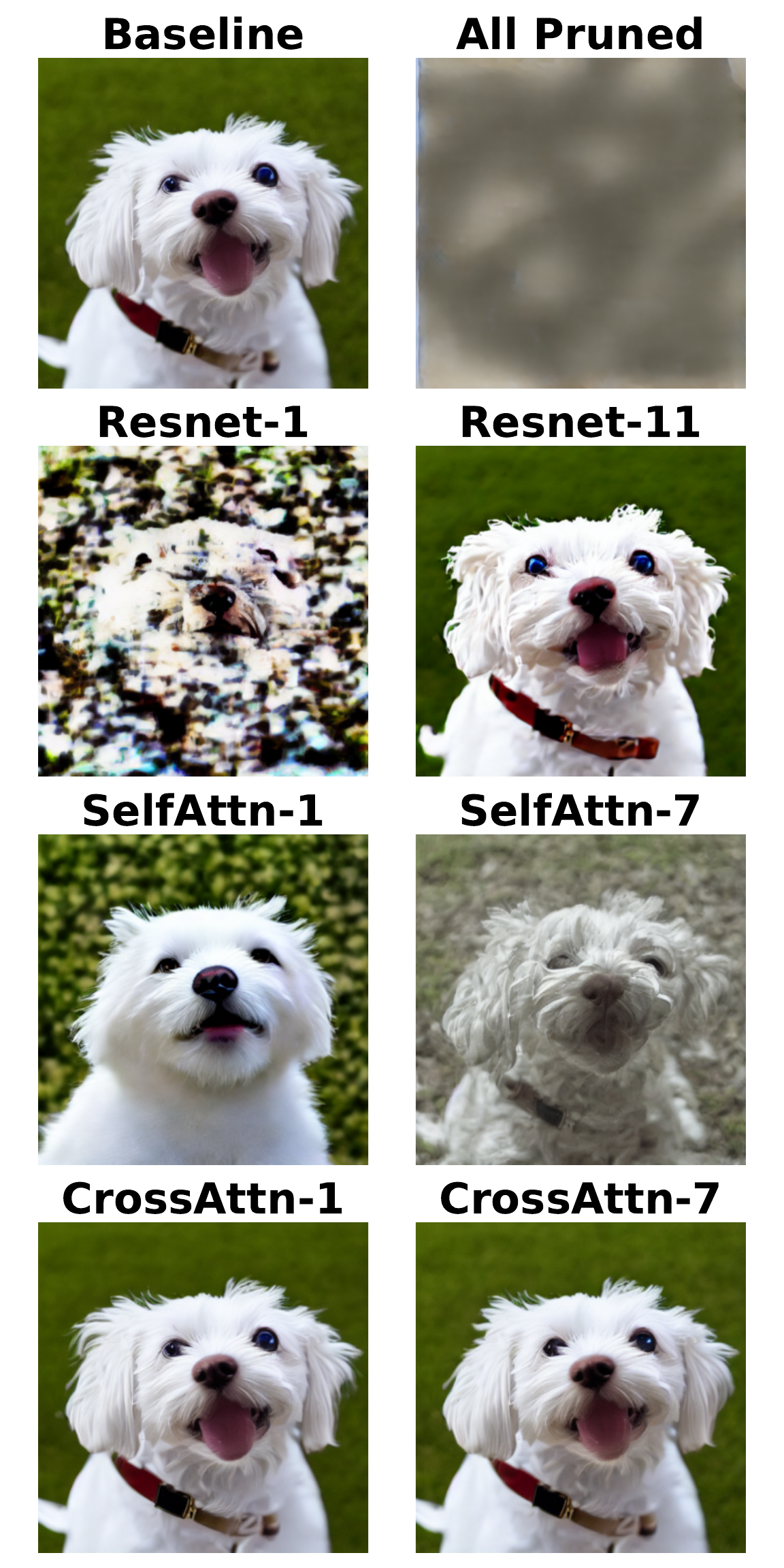}} 
    \subfigure[]{\includegraphics[width=0.66\columnwidth]{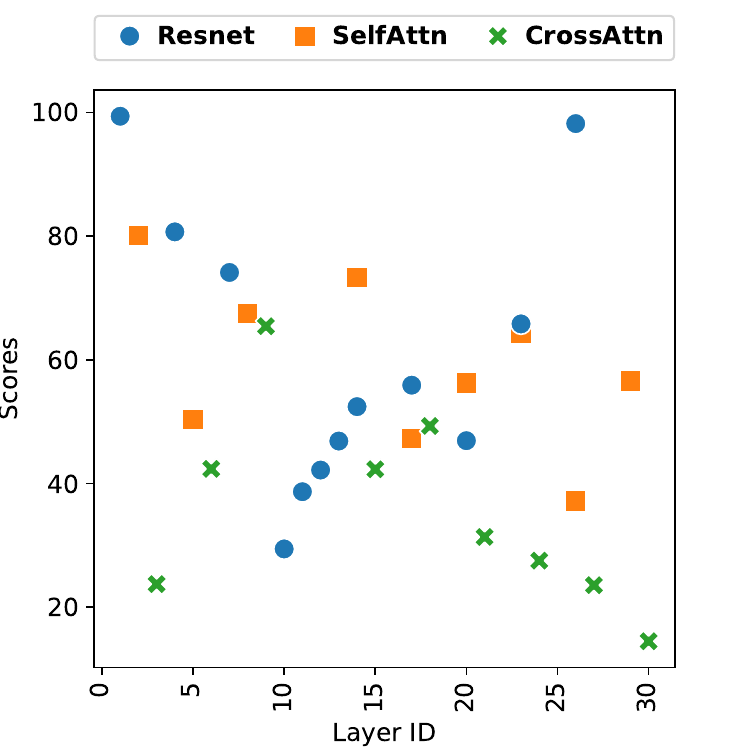}}
    \caption{(a) Different impact of pruning 50\% parameters in BK-SDM-Small without fine-tuning. (b) Evaluation score. the higher the more important.}
    \label{fig:analysis_prune}
    \vspace{-0.1in}
\end{figure}

We first investigate the filter redundancy in different components of U-Net, including ResNet, Self-Attention, and Cross-Attention blocks. By directly eliminating 50\% of parameters from these blocks, we assess the repercussions on the final image synthesis. Specifically, we apply the L1-Norm-based filter pruning for ResNet blocks, removing half of the filters accordingly. In attention-based blocks, we adopt a grouped L1-Norm approach to prune 50\% of the attention heads. Figure \ref{fig:analysis_prune}(a) illustrates the visual outcomes of this procedure. The experiments elucidate two primary insights: 1) The drastic reduction of parameters by half in some layers does not induce a conspicuous deterioration in the generated content's quality. Notably, the pruning of Cross-Attention blocks has a marginal effect on the output image, indicative of a high tolerance for parameter reduction. Similarly, the removal of parameters from the 11th ResNet block results in negligible changes, highlighting a substantial parameter redundancy within these particular layers; 2) Given the disparate contributions of individual layers to the holistic output quality, it is evident that a tailored, layer-specific pruning strategy is imperative. This underscores the necessity for variable pruning ratios to optimize performance while minimizing the loss of generative integrity.
In addition to subjective perception, we aim to measure the relative importance of each layer through objective indicators. We follow the idea in \cite{castells2024ldpruner} to calculate a significance evaluation score for each modification, which is defined as follows:
\begin{equation}
\label{eq2:score}
    score = \|avg(z_0) - avg(z_0')\|_2 + \|std(z_0) - std(z_0')\|_2
\end{equation}
\noindent where $z_0$ and $z_0'$ are original and modified latent representations, respectively. $avg({\cdot})$ is the average function, $std(\cdot)$ denotes the standard deviation, $\| \cdot \|2$ denotes the Euclidean norm. This score denotes the sensitivity of both shifts in the central tendency and changes in the variability of the latent representations. A higher score means that a modification is more significant to the model's performance. In practice, we use 50 different prompts to calculate the score. As depicted in Figure \ref{fig:analysis_prune}(b), the score of Cross-Attention blocks is generally lower than that of ResNet blocks and Self-Attention layers, which is consistent with the visualization result. The most important layer is the first ResNet block, pruning this layer without fine-tuning results in a severe performance drop in the final generated image.

\noindent\textbf{Pruning Procedure.}
Our objective is to swiftly generate a compact model that preserves the generative capability of the large model. To achieve this, we leverage a strategic pruning approach guided by the relative score outlined in Eq.\ref{eq2:score}. Specifically, we introduce two thresholds $a$ and $b$ to determine the pruning ratio. Layers deemed highly important, with a rank exceeding $b$, undergo mild pruning at a rate of 25\%. Conversely, those with significance below $a$ are more aggressively pruned at 75\%. Layers falling between these thresholds receive a moderate pruning ratio of 50\%. Upon attaining the targeted model sizes, we employ distillation strategies \cite{kim2023architectura} to fine-tune the small models to mimic the behavior of their larger counterparts:
\begin{equation}
    \mathcal{L}=\mathcal{L}_{\text {Task }}+\lambda_{\text {OutKD }} \mathcal{L}_{\text {OutKD }}+\lambda_{\text {FeatKD }} \mathcal{L}_{\text {FeatKD }} .
\end{equation}
where $\mathcal{L}_{\text {Task }}$ is the task loss (\ie MSE loss between added noise and actual noise),  $\mathcal{L}_{\text {OutKD}}$ denotes the output-level distillation (\ie MSE loss between outputs of each block in the U-net), and $\mathcal{L}_{\text {FeatKD}}$  denotes the feature-level distillation (\ie MSE loss between final output of teacher U-net and student U-net). $\lambda_{\text {OutKD}}$ and $\lambda_{\text {FeatKD}}$ are hyper-parameters controlling the weight of losses.

\begin{table*}[ht]\small
\caption{Results on zero-shot MS-COCO 2014 30K. 
For the hybrid results, only the small model's parameters are reported, as they can be deployed on the resource-constrained edge devices.}
\label{tab:t2i_coco}
\centering
\begin{tabular}{lccccc}
\toprule
Inference Method &   FID $\downarrow$ &    IS $\uparrow$ &  CLIP $\uparrow$ & \#Params (M) &FLOPs(T) $\downarrow$ \\
\midrule
Only SD-v1.4 & 12.22 & 37.63 &  0.2993 & 859.52 & 33.89 \\
Only BK-SDM-Small & 16.86 & 31.74 &  0.2692 & 482.34 &  21.78 \\
Only BK-SDM-Tiny & 17.05 & 30.37 &  0.2673 & 323.38 & 20.51 \\
Only OursTiny & 16.71 & 28.68 & 0.2611 &  224.49 &  15.14 \\



\midrule


Hybrid SD-v1.4 + Small ($k$ = 10) & 14.29 & 36.67 &  0.2921  & 482.34 &   26.62 \\
Hybrid SD-v1.4 + Tiny ($k$ = 10) & 14.59 & 35.70 &  0.2909 & 323.38 &   25.86 \\
Hybrid SD-v1.4 + OursTiny ($k$=10) & 13.75 & 34.49 & 0.2887 &  224.49 &   22.64 \\

\midrule
Hybrid SD-v1.4 + Small ($k$ = 5) & 15.48 & 34.02 &  0.2805 &  482.34 &   24.20 \\
Hybrid SD-v1.4 + Tiny ($k$ = 5) & 15.92 & 32.66 &  0.2780 &  323.38 &  23.19 \\
Hybrid SD-v1.4 + OursTiny ($k$=5) & 15.39 & 31.50 & 0.2734 &  224.49 & 18.89 \\



\midrule
Only SD-v1.4 LCM & 16.31 & 37.24 &  0.2825 & 859.60 & 10.86 \\
Only OursTiny LCM & 23.42 & 25.56 &  0.2309 & 224.57 &  4.85 \\
Hybrid SD-v1.4 LCM + OursTiny LCM (k=4) & 16.19 & 31.76 &  0.2698 & 224.57 & 7.86 \\
\bottomrule
\end{tabular}
\end{table*}

\noindent\textbf{Improved VAE.}
As the U-Net becomes smaller and the number of sampling iterations decreases, the VAE contributes more to the overall inference costs of the text-to-image generation pipeline. Despite TAESD~\footnote{\url{https://github.com/madebyollin/taesd}} designing a smaller VAE model to meet the demands of edge-side inference, there remains a significant gap in generation quality compared to the larger SD-v1.4 VAE.
Therefore, to enhance the decoding capabilities of VAE, we propose to train a lightweight VAE with advanced training strategies.
To match the latent space of the SD-v1.4 VAE, we distill our VAE encoder from it with L2 loss.
We train the VAE decoder as a standalone conditional model, leveraging a fixed VAE encoder to generate latent representations. These latent representations are then fed into the decoder. To optimize the decoder, we adopt the LPIPS loss~\cite{zhang2018unreasonable} and incorporate adversarial training to enhance the quality and detail of the generated images. Specifically, we leverage the projected discriminator from~\cite{sauer2023stylegan} but omit the conditional embeddings. We train our decoder with hinge loss.
We quantitatively compare our VAE to the origin SD-v1.4 VAE in Table \ref{tab:tiny_vae} and our VAE shows competitive reconstruction quality with exceeding parameter efficiency.

\section{Experiments}
\subsection{Experiment Settings}
\textbf{Base Models.} 
We present quantitative results for one large model, SD-v1.4 
 \cite{hf-sdv1-4}, as well as three smaller models including BK-SDM-Small, BK-SDM-Tiny~\cite{kim2023architectura} and OursTiny.  We use SD-v1.4 as the teacher model to fine-tune our pruned tiny model. We further leverage hybrid inference with acceleration methods LCM~\cite{luo2023latent}.
 For a qualitative assessment, we also display images generated by the Realistic Model \cite{Realistic_Vision_V5} and its according tiny model segmind small sd \cite{segmind_small}. Qualitative results can be seen in Appendix \ref{qualiative realistic}.

\noindent\textbf{Evaluation and Datasets}
We use 30K prompts from the zero-shot MS-COCO 2014~ \cite{lin2014microsoft} validation split and compare the generated images to the whole validation set. Frechet Inception Distance (FID) \cite{heusel2018gans} and Inception Score (IS) \cite{salimans2016improved} are adopted to assess visual quality. CLIP score \cite{hessel2022clipscore} with CLIP-ViT-g/14 model is also reported to assess text-image correspondence. We adopt the widely-used protocols, \ie the number of parameters and required Float Points Operations (denoted as FLOPs). The smaller models produced by the proposed pruning method are trained on a subset of 0.22M image-text pairs from LAION-Aesthetics V2 6.5+, which represents less than 0.1\% of the training pairs used in the LAION-Aesthetics V2 5+\cite{schuhmann2022laion5b} for training SD-v1.4.

\noindent\textbf{Implementation details.} We adjust the code in Diffusers \cite{von-platen-etal-2022-diffusers} for hybrid inference pipeline and distillation. A single NVIDIA A100 80G GPU is used for training small models. For compute efficiency, we always opt for 25-step DPM-Solver~\cite{lu2022dpm} at the inference phase, unless specified. For LCM models, we adopt an 8-step sampling. The classifier-free guidance scale is set to 7 by default. The latent resolution is set to the default, yielding 512×512 images. For a fair comparison, we
follow BK-SDM~\cite{kim2023architectura} to resize generated images to 256x256 and calculate FID, IS and CLIP score.



\subsection{Quantitative Results.}
Table \ref{tab:t2i_coco} shows the quantitative results on 30K samples from the MS-COCO 2014 30K validation set.

\noindent\textbf{Advantages of Smaller Models.} 
 The OursTiny model generated by the proposed pruning algorithm, significantly reduces parameter count (224.49M), in comparison to BK-SDM-Tiny (323.38M) and BK-SDM-Small (482.34M), with an FID (16.71) that remains close, indicating minimal compromise on the quality of generated images. With FLOPs at 15.14T, OursTiny demonstrates a significant saving in computational cost, much lower than other compact models. This is particularly crucial for resource-constrained edge devices, enhancing deployment efficiency and energy efficiency.



\begin{table}[ht]
\caption{Comparison between SD-v1.4 VAE, TAESD, and our lightweight VAE in terms of parameters, latency (ms, on V100 GPU), and FID scores (on MS-COCO 2017 5K). We omit CLIP scores in the reconstruction evaluations.
Additionally, we compare SD-v1.4 VAE, TAESD, and our VAE deployed with the LCM models using MS-COCO 2014 30K prompts for text-to-image tasks.}
\label{tab:tiny_vae}
\setlength\tabcolsep{3pt} 
\setlength\extrarowheight{1.5pt}
\centering
\scalebox{0.7}
{
\begin{tabular}{llcccccc}
\toprule
~ & Inference Method  & FID $\downarrow$ & CLIP $\uparrow$ & \#Params (M) & Latency (ms) \\
\hline
\multicolumn{1}{l|}{\multirow{3}{*}{Only VAE}} & SD-v1.4 VAE    & 3.60 & - &    83.7 & 427.2\\
\multicolumn{1}{l|}{\multirow{3}{*}{Reconstructions}} &  TAESD  & 6.84 & - &   2.4  & 30.7\\
\multicolumn{1}{l|}{}  &        Ours VAE       & 5.47 & - &    2.4  & 30.7\\
\hline
\multicolumn{1}{l|}{\multirow{8}{*}{VAE with}}& SD-v1.4 LCM    & 16.30 & 0.2825 &  909.0 & 1733.6\\
\multicolumn{1}{l|}{\multirow{8}{*}{Text-to-Image}} &   + TAESD   & 15.26 & 0.2811 &   861.9  & 1337.1\\
\multicolumn{1}{l|}{\multirow{8}{*}{LCM Models}}&   + Ours VAE     & 15.62 & 
0.2814 &   861.9  & 1337.1\\
\cline{2-6} 
\multicolumn{1}{l|}{}&    OursTiny LCM   & 23.42 & 0.2309 &  274.1 & 1145.8\\
\multicolumn{1}{l|}{} &   + TAESD        & 23.19 & 0.2294 &  225.8  & 749.3\\
\multicolumn{1}{l|}{} &   + Ours VAE     & 23.06 &  0.2298 &  225.8  & 749.3\\
\cline{2-6} 
\multicolumn{1}{l|}{}&  OursTiny LCM (k=4)  & 16.19 & 0.2698 &  274.1 & 1439.7 \\
\multicolumn{1}{l|}{} &   + TAESD        & 15.82 & 0.2683 &  225.8 & 1043.2 \\
\multicolumn{1}{l|}{} &   + Ours VAE     & 15.79 & 0.2687 &  225.8 & 1043.2 \\
\hline
\end{tabular}
}
\end{table}
\noindent\textbf{Advantages of Hybrid SD.}
Hybrid strategies based on SD-v1.4, when combined with BK-SDM-Small, BK-SDM-Tiny, and OursTiny, generally show better performance in FID and IS.
For example, \textit{Hybrid SD-v1.4 + OursTiny (k=10)} manage to significantly reduce FID to 13.75, compared to \textit{BK-SDM-Tiny} with an FID of 17.05. The hybrid models have comparable CLIP scores to \textit{SD-v1.4}, ensuring similar capabilities in semantic alignment between generated images and text prompts. The hybrid models offer flexibility through the tunable parameter $k$. For instance, comparing \textit{Hybrid SD-v1.4 + Tiny (k=10)} with \textit{Hybrid SD-v1.4 + Tiny (k=5)},  we see that reducing $k$ can be a trade-off strategy. While FID increases from 14.59 to 15.92, this adjustment could be beneficial in scenarios where computational constraints are tighter, as FLOPs decrease from 25.86T to 23.19T, indicating a more computationally efficient setup at the expense of slightly reduced image fidelity. Moreover, the hybrid models offer flexibility through the tunable parameter $k$. Comparing \textit{Hybrid SD-v1.4 + Tiny (k=10)} with \textit{Hybrid SD-v1.4 + Tiny (k=5)}, we see that reducing $k$ can be a trade-off strategy. While FID increases from 14.59 to 15.92, this adjustment could be beneficial in scenarios where computational constraints are tighter, as FLOPs decrease from 25.86T to 23.19T, indicating a more computationally efficient setup at the expense of slightly reduced image fidelity. 

\noindent\textbf{Hybrid SD with Acceleration Methods.} In this part, we explore the flexibility of  Hybrid SD when integrating with diffusion acceleration methods. We leverage the popular step-distillation acceleration method LCM, which maps data from noise directly through consistency distillation and improves the sample quality by alternating denoising and noise injection during inference. 
For a fair comparison, all models are trained with the exact same training setup. 
As depicted in Table \ref{tab:t2i_coco}, Hybrid SD shows good compatibility with LCM. Hybrid SD surpasses the small model in both sample quality (FID: 16.19 v.s. 23.42) and text-image alignment (CLIP: 0.2698 v.s. 0.2309).

\begin{figure}[tb]
    \centering
    \includegraphics[width=1.0\linewidth]{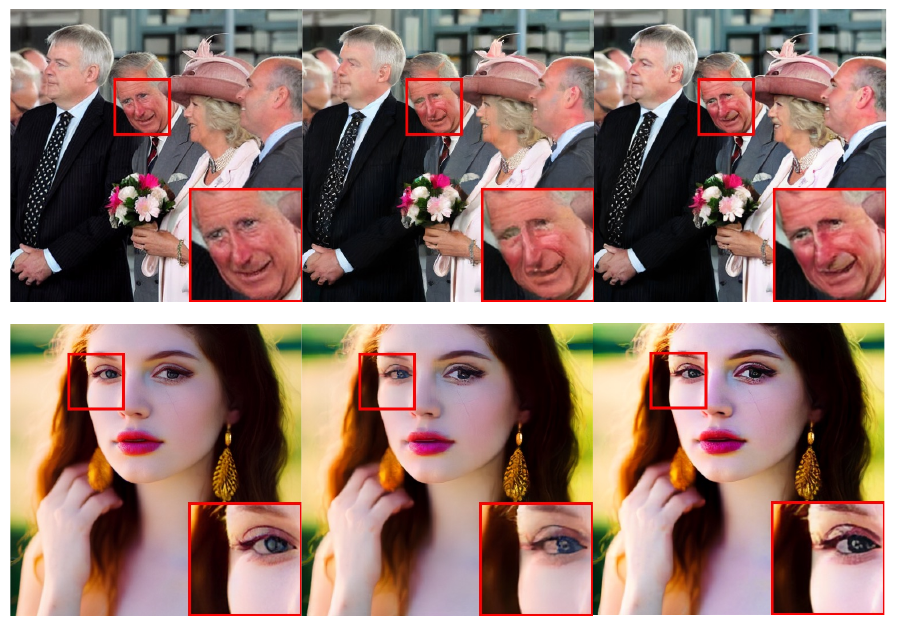}
    \text{SD-v1.4 \qquad\qquad TAESD \qquad\qquad Ours VAE}
    \caption{Visualizations of images generated by SD-v1.4 VAE (left), TAESD (middle), and ours VAE(right). The first row shows images reconstructed directly by VAE while the second row denotes images decoded from the latent generated by SD-v1.4 LCM. Our VAE shows competitive performance compared to SD-v1.4 VAE while excelling TAESD in terms of detail generation and color saturation.}
    \label{fig:vae_visual}
    \vspace{-0.05in}
\end{figure}

\noindent\textbf{Advantages of our lightweight VAE.} We comprehensively benchmark our lightweight VAE against the original SD-v1.4 VAE and the open-source TAESD. We evaluated the above VAE in both reconstruction tasks and text-to-image generation tasks. For reconstruction tasks, we calculate the FID score on MS-COCO 2017 5K dataset. Ours VAE performs better than TAESD (FID: 5.47 v.s. 6.84) in visual quality while enjoying superior parameter efficiency than SD-v1.4 VAE (2.4M v.s. 83.7M). We also evaluate our VAE on text-to-image tasks with LCM models on MS-COCO 2014 30K. Our VAE enjoys a huge reduction in latency and the number of parameters. Our VAE shows a better FID even than the original SD-v1.4 VAE, and a slightly minor drop in CLIP. 
We provide visualizations in Figure \ref{fig:vae_visual},  demonstrating our VAE excels TAESD in terms of detail generation and color saturation.



\vspace{-0.1in}
\subsection{Qualitative Results.}
\begin{figure*}[ht]
    \centering
    \includegraphics[width=1.0\linewidth]{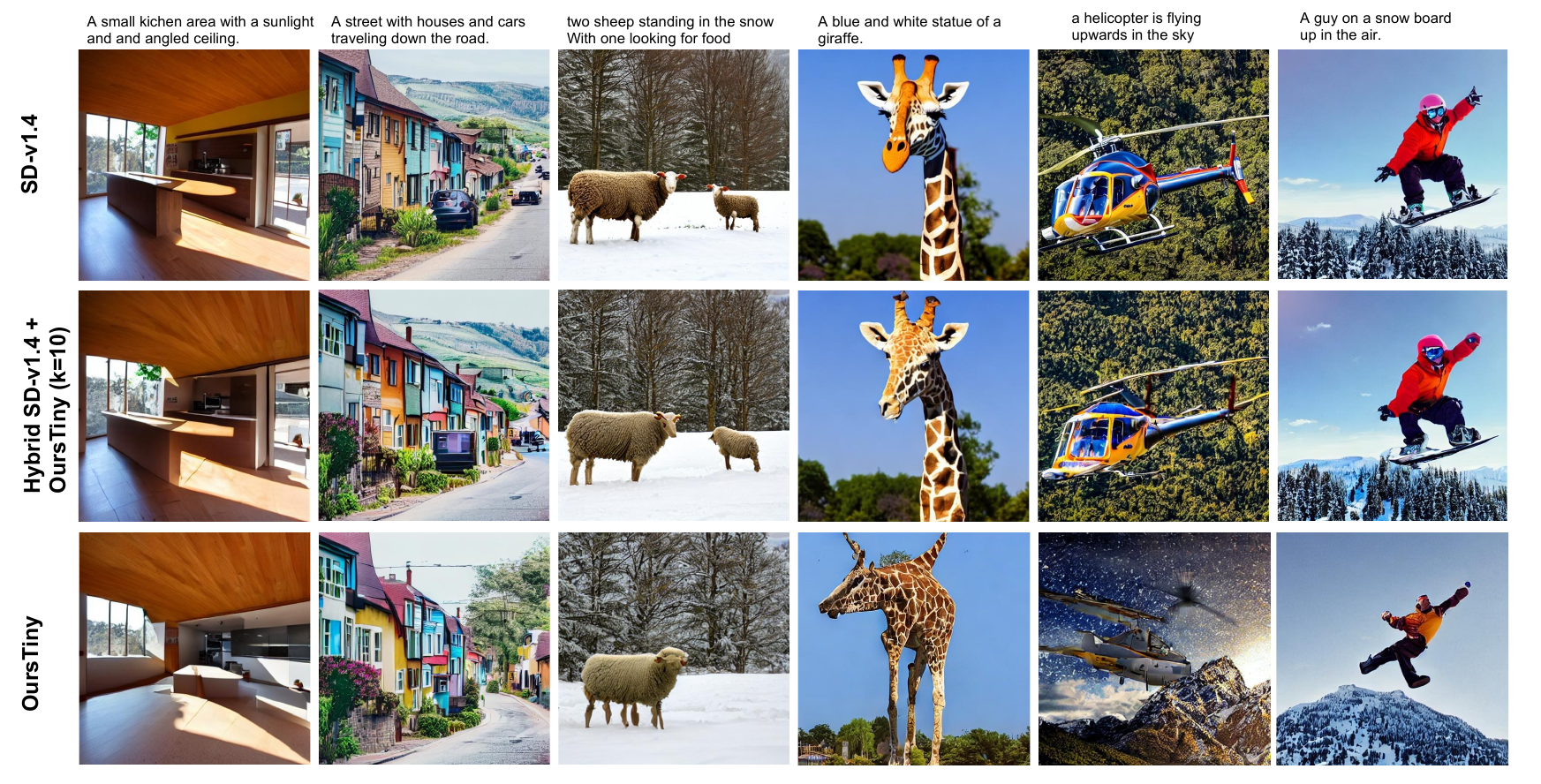} 
    \caption{Visualization of the generated images. We use prompts in MS-COCO 2017 5K validation set. Some prompts are omitted from this section for brevity. While the smaller model exhibits a slight degradation in semantic detail compared to SD-v1.4, our Hybrid SD adeptly maintains semantic consistency.}
    \label{fig:hybrid_sdv14_small}
    \vspace{-0.01in}
\end{figure*}
The primary advantage of our Hybrid SD approach lies in its ability to preserve the semantic information typically associated with larger models. This feature is visibly evident in the resulting visual outputs.

\noindent\textbf{Results on Basic models.}
 As evident from the showcased examples in Figure \ref{fig:hybrid_sdv14_small} the images generated by our method exhibit a greater consistency than those generated by the large diffusion model. 
 The capacity of the small model to incorporate textual cues into image synthesis is notably inferior to its larger counterpart. This is exemplified in instances where the small model fails to comprehend errors or specific details – like the misspelled ``\textit{kichen}" and the disjointed reference to a ``\textit{snow board}". Furthermore, the small model occasionally struggles with straightforward prompts, as seen in the inability to generate an image depicting ``\textit{two sheep}", thereby accentuating the disparity in text-to-image translation proficiency between models of differing sizes. We also provide qualitative results of realistic model~\cite{Realistic_Vision_V5}. Please refer to Appendix \ref{qualiative realistic}.

\begin{table}[ht]
\caption{Evaluation of LCM models with cloud inference, edge inference, and our edge-cloud collaborative inference. The FLOPs column represents the computational overhead on the cloud, while the numbers in $(\cdot)$ indicate the computations on the edge devices.  * means inference with our VAE.}
\setlength\extrarowheight{1.2pt}
\scalebox{0.75}
{
\begin{tabular}{lllcc}
\hline
\multicolumn{1}{l|}{} & {Inference Method}   & Latency (ms) & FLOPs (T)
\\ \hline
\multicolumn{1}{l|}{\multirow{3}{*}{Only Cloud}}                          & SD-v1.4 LCM      &    1733.6  &  15.8 (+0)  \\
\multicolumn{1}{l|}{\multirow{3}{*}{(V100 GPU)}}       &  OursTiny LCM         &  1080.4 ($\downarrow$ 38\%)   &  9.8(+0) \\
\multicolumn{1}{l|}{}  &  OursTiny LCM (k=4)  &  1407.0 ($\downarrow$ 19\%)   &   12.8 (+0)  \\
\multicolumn{1}{l|}{}       &  OursTiny LCM* (k=4)   &   1010.5 ($\downarrow$ 42\%)   & 8.1 (+0)     \\ \hline
\multicolumn{1}{l|}{\multirow{3}{*}{Only Edge}}                             &  SD-v1.4 LCM     & 3959.6  & 0 (+15.8) \\
\multicolumn{1}{l|}{\multirow{3}{*}{(iPhone 15 Pro)}}        & OursTiny LCM     &  2340.4 ($\downarrow$ 41\%)    & 0 (+9.8)    \\
\multicolumn{1}{l|}{}      &  OursTiny LCM (k=4)  &  3150.0 ($\downarrow$ 20\%)  &  0 (+12.8) \\
\multicolumn{1}{l|}{}      &  OursTiny LCM* (k=4) &  2799.0 ($\downarrow$ 29\%)  &  0 (+8.1) \\ \hline
\multicolumn{1}{l|}{\multirow{1}{*}{Edge-Cloud}}                            &  OursTiny LCM (k=4)    &    2004.3   &  5.4 (+7.4) \\
\multicolumn{1}{l|}{\multirow{1}{*}{Collaborative}}                         &  OursTiny LCM* (k=4)   &   1653.3    &  5.4 (+2.7) \\
\hline
\label{edge_cloud}
\end{tabular}
}
\vspace{-2em}
\end{table}

\subsection{Edge-Cloud Collaborative Inference}

\noindent Table \ref{edge_cloud} presents a comparison of FLOPs and latency between only cloud, only edge, and edge-cloud collaborative inference. We adopt LCM Models with a default 8-step sampling on a V100 GPU for cloud inference and on an iPhone 15 Pro for edge inference. As depicted in Table \ref{edge_cloud}, our hybrid inference strategy achieves substantial reductions in FLOPs and latency across all three deployments compared to the original large model inference. 
Furthermore, our VAE consistently demonstrates efficiency gains. 
Our proposed hybrid inference achieves a $49\% $ reduction in FLOPs (8.1 T v.s. 15.8T) and a corresponding $42\%$ decrease in Latency (1010.5 ms v.s. 1733.6 ms) on cloud servers. By further leveraging edge-cloud collaborative inference, we successfully offload 2.7 T FLOPs to the edge devices, reducing a cost of 66\% in cloud servers (5.4 T v.s. 15.8 T).
It is noteworthy that the edge-cloud collaborative inference has lower FLOPs while exhibiting higher latency than the only cloud inference. This is due to the relatively lower capabilities of edge GPUs compared to high-end cloud GPUs. However, the minor increase in latency is an acceptable trade-off for the significant cost reductions achieved by offloading computations to edge devices. Moreover, as edge GPU continues to improve, the benefits of our hybrid inference will be further amplified. 
\vspace{-0.03in}

\vspace{-0.01in}

\section {Conclusion}
In conclusion, we introduce Hybrid SD,  a novel edge-cloud collaborative inference framework designed to enhance cost-effectiveness by seamlessly integrating cloud and edge capabilities for diffusion model inference. 
We further prune the SDMs U-Net and train a lightweight VAE, achieving state-of-the-art parameter efficiency on edge devices. Extensive experiments demonstrate our approach can reduce the cloud cost by $66\%$ with competitive visual quality. We also deploy Hybrid SD with acceleration methods, showing its superior compatibility.
Our findings lay the groundwork for expanding the scope of hybrid inference strategies to broader application areas and refining efficiency through additional optimization techniques. We anticipate that this study will spur innovative research endeavors aimed at advancing the practical implementation and scalability of diffusion models in hybrid inference contexts. \\
\textbf{Limitations.} While our approach can deploy smaller diffusion models on devices with semantic-preserving hybrid inference, the number of parameters is still relatively large. A possible solution is to combine quantization strategies to further compress the models, which we leave as future work.

\newpage
\appendix

\nobibliography*

\bibliography{aaai25}

\newpage

\onecolumn

\setcounter{secnumdepth}{1}

\section{Algorithm}\label{alg: hybrid}
\begin{algorithm}[htb] 
   \caption{Hybrid Inference}
\begin{algorithmic}
\STATE{\bfseries Input:} large model deployed on the cloud service $\vtheta_{large}$, small model deployed on the edge device $\vtheta_{small}$, split step $k$, condtioning $\vc$, step $t$, total inference step $T$, noise schedule $\beta_t$, decoder $\mathcal{D}(\cdot)$, ODE Solver $\Phi(\cdot, \cdot, \cdot, \cdot,\cdot)$. \\

\STATE Sample $\vz_T\sim \mathcal{N}(0, \mathbf{I})$
$t \leftarrow T$
\WHILE{$t>k$}
 \STATE $\vz_{t-1} = \Phi(\vtheta_{large}, \vz_t, c, t, t-1)$ 
 \STATE $t \leftarrow t -1$
\ENDWHILE
\STATE ($\vz_{t}, t, c$) is sent to edge devices, and switch to inference with small model on edge
\WHILE{$t > 0$}
 \STATE$\vz_{t-1} = \Phi(\vtheta_{small}, \vz_t, c, t, t-1)$ 
 \STATE $t \leftarrow t -1$
 \ENDWHILE
\STATE $\vx \leftarrow \mathcal{D}(\vz_0)$
\STATE{\bfseries Output:} $\vx$
\end{algorithmic}
\end{algorithm}

\section{Analysis of environmental impact.}
The advancement of generative AI has also ushered in considerations regarding its environmental impact \cite{berthelot2024estimating, wu2022sustainable}. \cite{berthelot2024estimating} measured that the energy consumed by a single inference of SD1.5 is $1.38\times10^{-3}$ kWh. The model inference on the cloud emits about 180 tons of carbon dioxide per year (equivalent to the carbon emissions of one person's life for 30 years \cite{strubell2019energy}) and consumes 1.24 GW of energy. Edge devices can inference with SD model with lower energy consumption, while also helping cloud service providers reduce the energy consumption of data centers, achieving environmental and sustainable development goals.

\section{More Qualitative Results.}
\label{qualiative realistic}



\begin{figure}[H]
    \centering
    \includegraphics[width=1.0\linewidth]{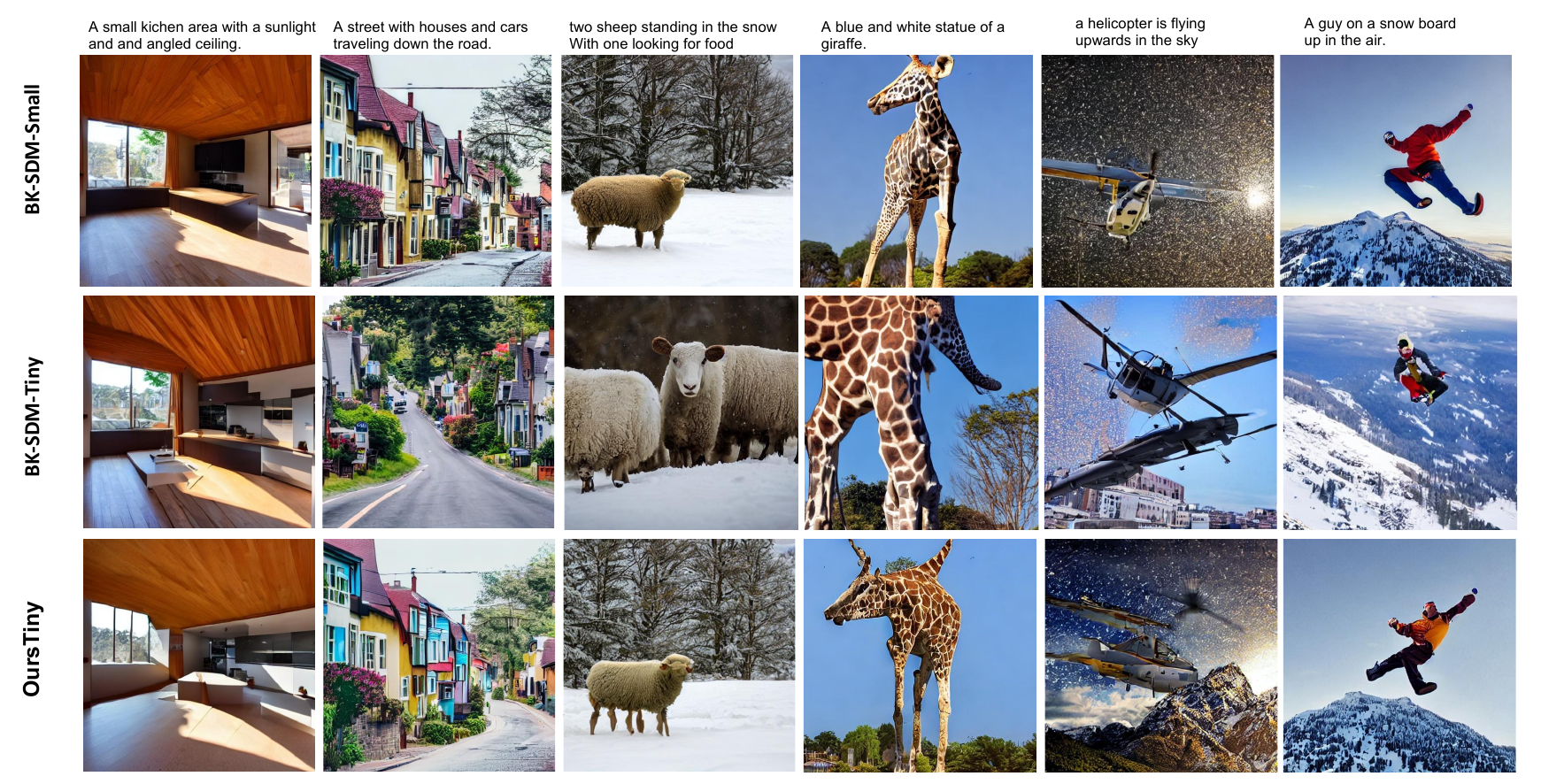} 
    \caption{More results on small models.}
    \label{fig:small-models}
\end{figure}

\noindent\textbf{Results on Small Models}
We provide more qualitative results of OursTiny in Figure \ref{fig:small-models}. Our smallest pruned OursTiny (224M) shows competitive quality compared to the larger  BK-SDM-Small (482M) and BK-SDM-Tiny (323M) with much smaller parameters.

\begin{figure}[h]
    \centering
    \includegraphics[width=0.72\linewidth]{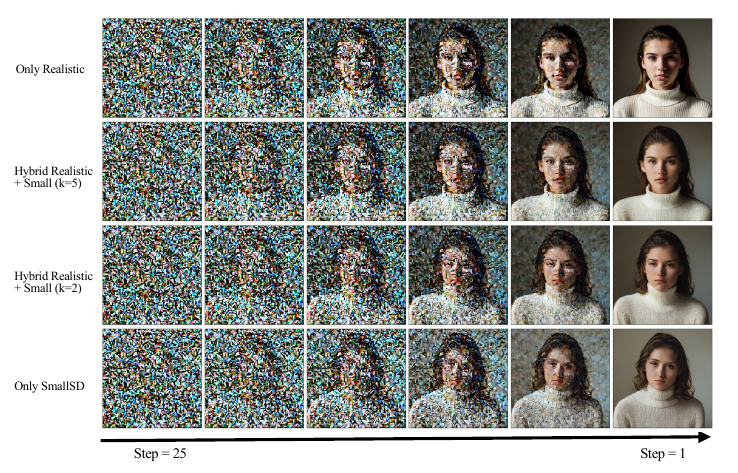} 
    \caption{Generated samples from different hybrid configurations given the same initial noise and text \textit{``Faceshot Portrait of pretty young (18-year-old) Caucasian wearing a high neck sweater"}.}
    \label{fig:realistic-image-0}
\end{figure}

\begin{figure}[h]
    \centering
    \includegraphics[width=0.89\linewidth]{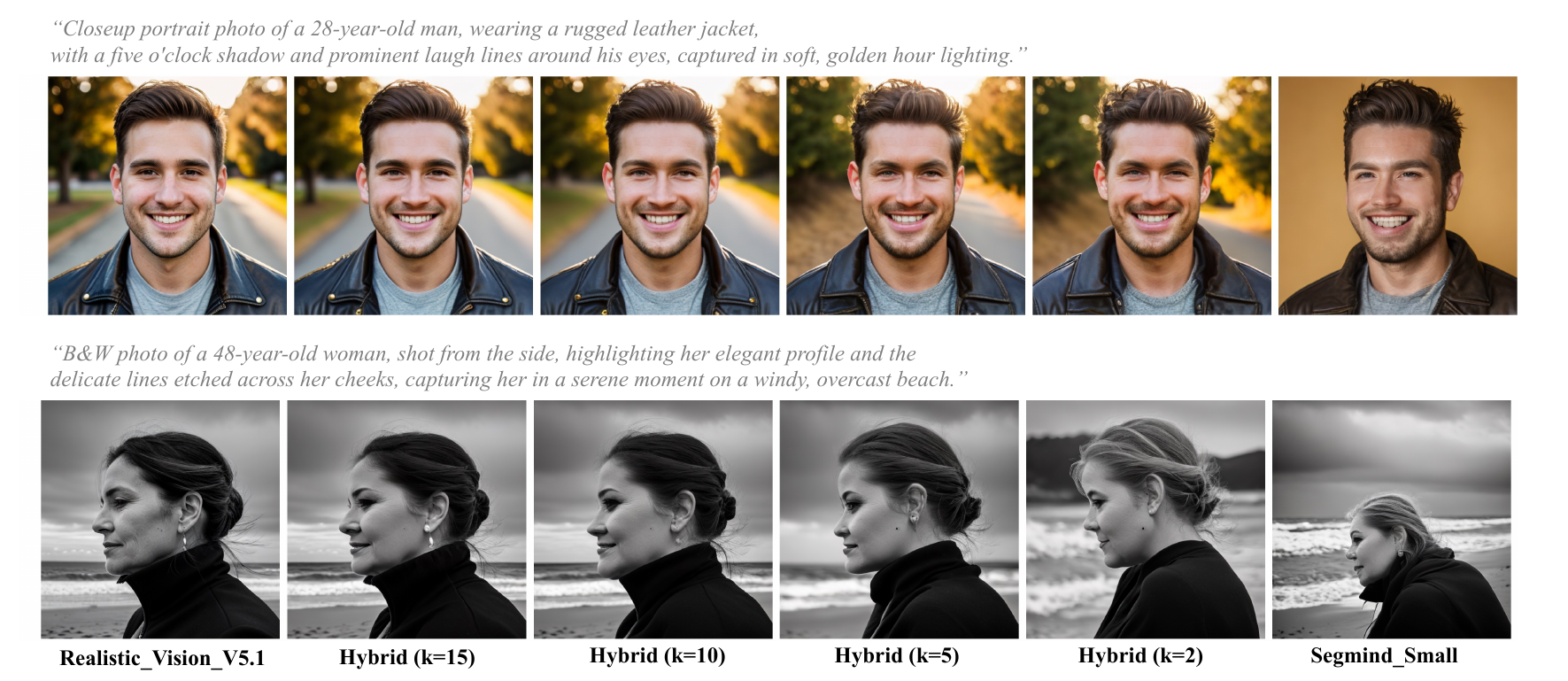} 
    \caption{More results on realistic models by using different split steps $k$.}
    \label{fig:realistic-image-1}
\end{figure}

\noindent\textbf{Results on Realistic Models.} 
We show more results on realistic models. Figure \ref{fig:realistic-image-0} illustrates the evolution from the preliminary stages to the final output. This visual narrative underscores the model's capability to refine details progressively. Notably, the small model struggles to incorporate specific directives, such as the ``18-year-old" attribute mentioned in the text prompt. Conversely, as the larger model undertakes additional inference steps, its impact becomes increasingly pronounced. A pivotal observation emerges when the large model executes 5 steps, under the hybrid inference paradigm, where the small model takes over for the remaining steps, yielding an image virtually indistinguishable from the large model's output. More illustrations of this enhanced realism are showcased in Figure \ref{fig:realistic-image-1}.

\begin{figure}[h]
    \centering
    \includegraphics[width=1.0\linewidth]{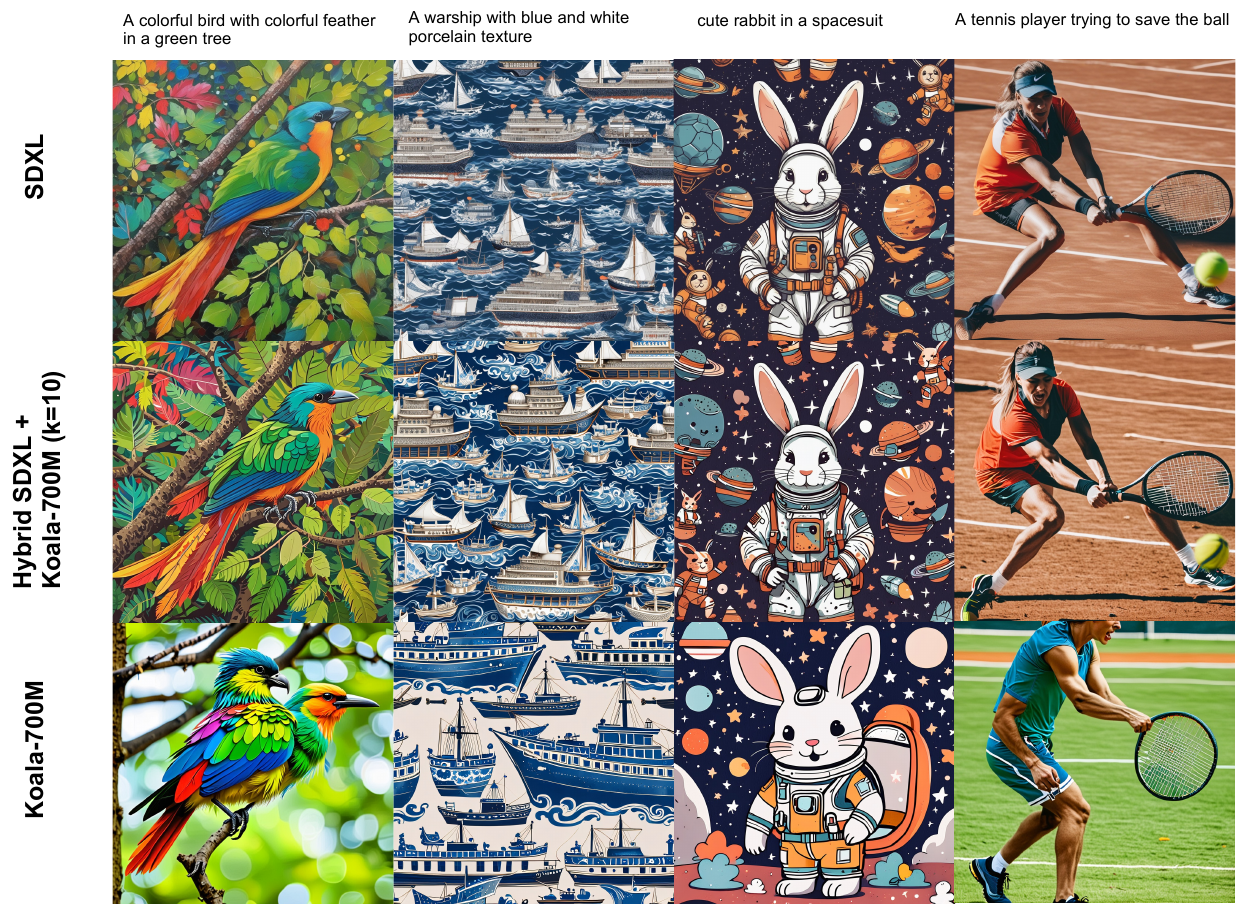} 
    \caption{Visualization of images generated by Hybrid SD with large model SDXL and smaller model Koala-700M.}
    \label{fig:sdxl_visual}
\end{figure}

\newpage
\noindent\textbf{Results on SDXL.} We further provide results of Hybrid SD on SDXL models in Figure \ref{fig:sdxl_visual}. We adopt the SDXL-base~\cite{podell2023sdxl} and its accordingly small model koala-700M~\cite{Lee@koala}. As depicted in Figure \ref{fig:sdxl_visual}, the small model underperforms the original SDXL in semantic planning and text-image alignment. For instance, given the prompt ``A tennis player trying to save the ball.", the small model generates a tennis player with a weird third arm and distorted tennis racket. In contrast, the large SDXL generates a much better image with natural semantic planning. Our hybrid inference shows good consistency with the large model's outputs while shifting 15 of 25 steps from cloud servers to edge devices.

\begin{figure}[h]
    \centering
    \includegraphics[width=1.0\linewidth]{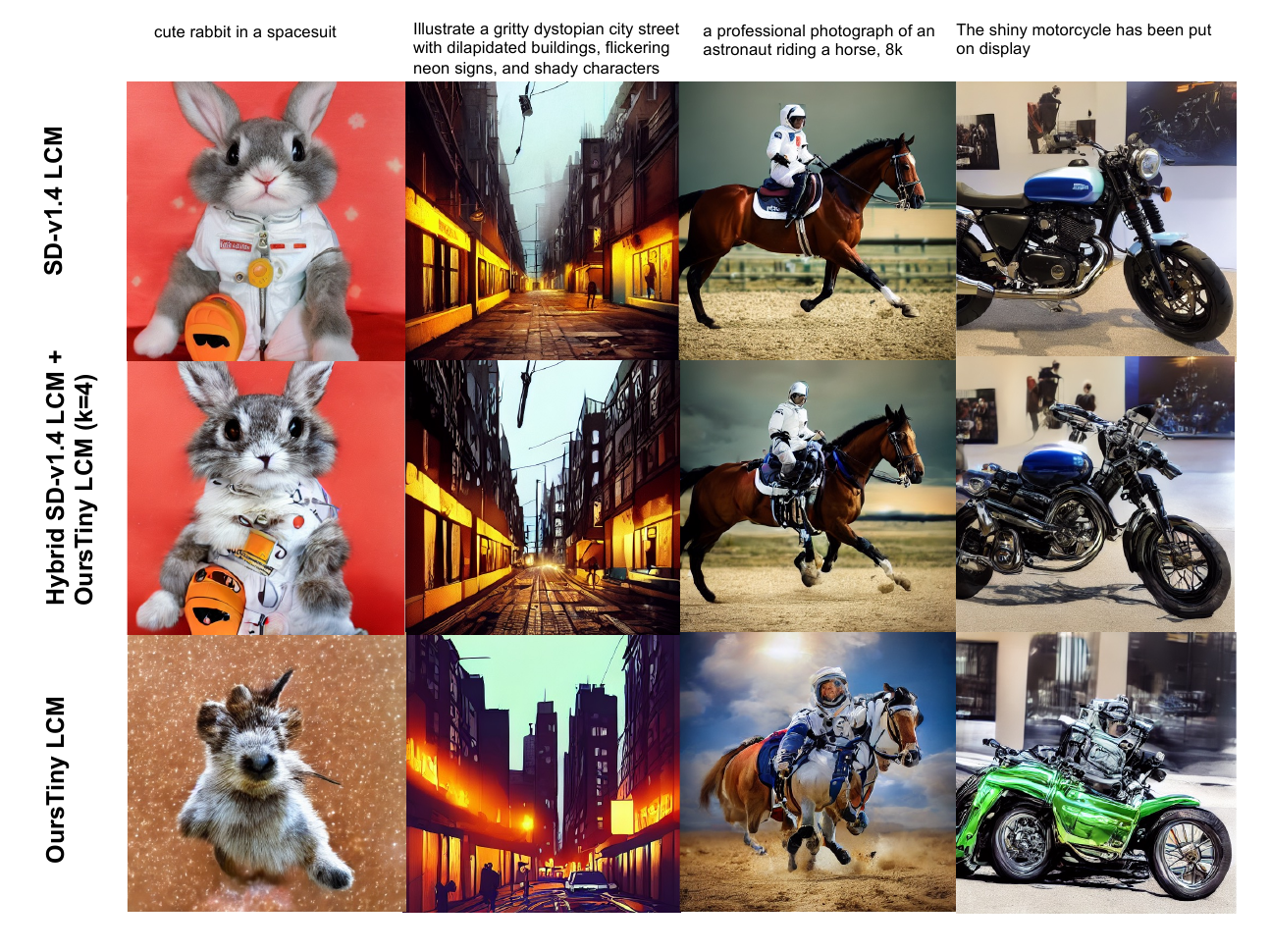} 
    \caption{Visualization of images generated by Hybrid LCM models with large model SD-v1.4 LCM and smaller model OursTiny LCM.}
    \label{fig:lcm_visual}
\end{figure}

\newpage
\noindent\textbf{Results on LCM Models.} We further provide results of hybrid inference on accelerated LCM models in Figure \ref{fig:lcm_visual}. The smaller LCM models fail to generate images with good semantics, due to the limitation of model size and model capability. For instance, given the prompt "The shiny motorcycle has been put
on display", the smaller models output the motorcycle with incomplete structures. The SD-v1.4 LCM models excel in visual planning and text-image alignment. Our Hybrid SD in the second row shows good preservation and consistency with the larger model while achieving much reduction in FLOPs and parameters.



\end{document}